\ifdefined\pdfcompresslevel
\fi
\ifdefined\pdfobjcompresslevel
\fi
\documentclass[sigconf]{acmart}

\AtBeginDocument{%
  }

\copyrightyear{2026}
\acmYear{2026}
\setcopyright{cc}
\setcctype{by}
\acmConference[MM '26]{Proceedings of the 34th ACM International Conference on Multimedia}{November 10--14, 2026}{Rio de Janeiro, Brazil}
\acmBooktitle{Proceedings of the 34th ACM International Conference on Multimedia (MM '26), November 10--14, 2026, Rio de Janeiro, Brazil}
\acmDOI{10.1145/3767308.3834966}
\acmISBN{979-8-4007-2213-4/2026/11}

\usepackage{amsmath} 
\usepackage{multirow}

\graphicspath{{./}}

\usepackage{makecell} 

\usepackage{pifont}
\newcommand{\cmark}{\ding{51}}

\begin{document}

\title{ProjFormer: Point Cloud Completion via Geometric-Projective Transformer and Cross-Modal Semantic Constraints}












\author{Sheng Liu}
\affiliation{%
  \institution{Hunan University}
  \city{Chang Sha}
  \country{China}
}
\email{liusheng23@hnu.edu.cn}

\author{Meng Wang}
\authornote{Corresponding author.}
\affiliation{%
  \institution{Hunan University}
  \city{Chang Sha}
  \country{China}
}
\email{willem@hnu.edu.cn}

\author{Ruihui Li}
\affiliation{%
  \institution{Hunan University}
  \city{Chang Sha}
  \country{China}
}
\email{liruihui@hnu.edu.cn}

\author{Huilong Pi}
\affiliation{%
  \institution{Hunan University}
  \city{Chang Sha}
  \country{China}
}
\email{phl880217@hnu.edu.cn}

\author{Zhuo Tang}
\affiliation{%
  \institution{Hunan University}
  \city{Chang Sha}
  \country{China}
}
\email{ztang@hnu.edu.cn}

\author{Kenli Li}
\affiliation{%
  \institution{Hunan University}
  \city{Chang Sha}
  \country{China}
}
\email{lkl@hnu.edu.cn}

\renewcommand{\shortauthors}{Sheng Liu et al.}

\begin{abstract}
Point cloud completion is inherently ill-posed due to severe sparsity and ambiguity in partial observations. Existing multi-view methods alleviate this by incorporating 2D semantics, but often rely on learned attention and fixed fusion, which lack geometric consistency and adaptability.
We propose ProjFormer, a cross modal framework that enforces geometry-consistent 2D--3D interaction through explicit projection and adaptive feature routing. A Projective Guided View Attention module aligns 3D points with multi-view features via deterministic projection, enabling efficient and geometrically consistent aggregation. Building on this, a geometry-aware routing network performs point-wise adaptive fusion of structural and observation-driven features for progressive refinement.
Experiments show that, under a lightweight design, ProjFormer delivers competitive performance with improved structural completeness.
\end{abstract}



\begin{teaserfigure}
  \includegraphics[width=\textwidth]{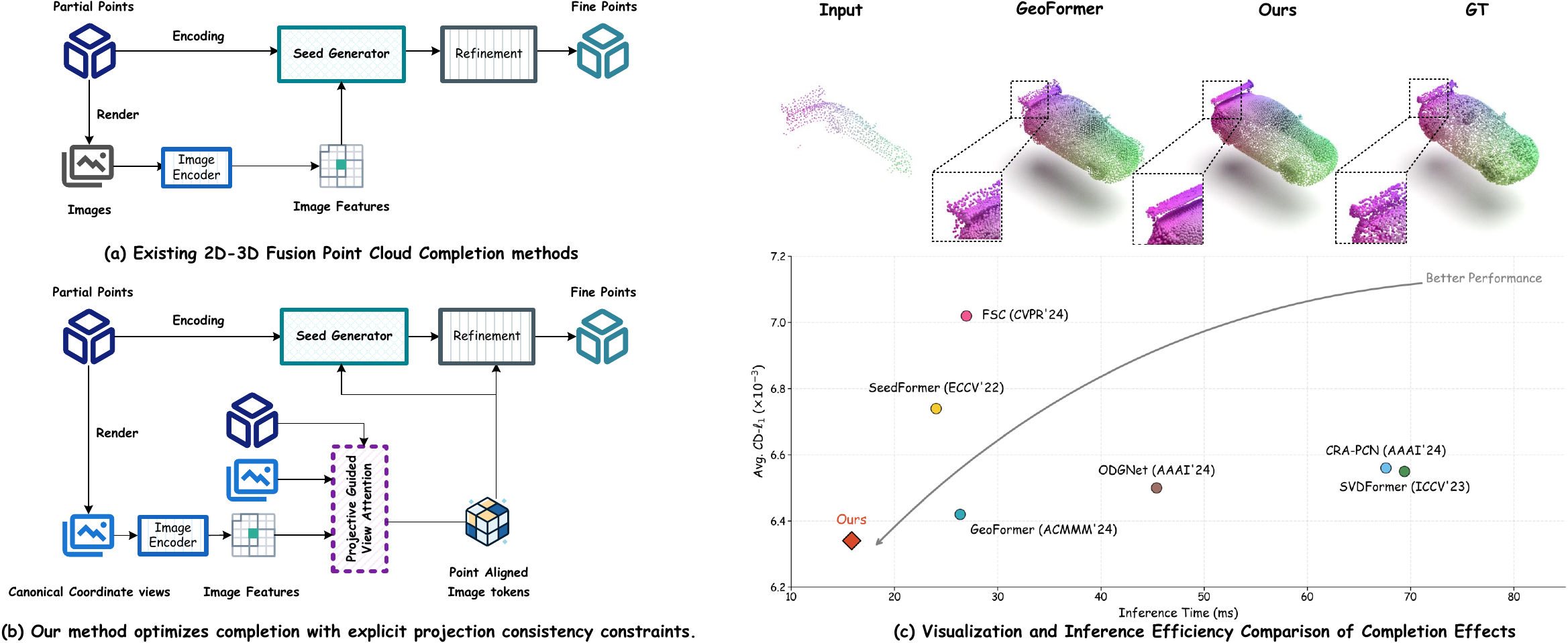}
  \caption{
  Unlike global 2D–3D fusion, ProjFormer deterministically aligns multi-view features with 3D points, enabling geometry-aware refinement with fewer artifacts and lower latency.
  }
  \Description{Comparison between existing 2D--3D fusion methods and our ProjFormer. Existing methods use global pooling, while ours uses deterministic projection for fine-grained semantic injection.}
  \label{fig:teaser}
\end{teaserfigure}

\maketitle

\section{Introduction}

3D point clouds have emerged as a fundamental representation in modern computer vision, underpinning a wide range of applications including autonomous driving, robotic navigation, and augmented reality.  
Unlike 2D images, point clouds explicitly represent 3D geometry without collapsing spatial structures through perspective projection.
To this end, early studies have proposed various deep learning architectures specifically designed for point clouds, such as the MLP-based PointNet and its variants\cite{qi2017pointnet,qi2017pointnet++}, as well as graph convolution-based frameworks\cite{wang2019dynamic}. However, due to sensor limitations, object occlusions, and sampling sparsity, real-world point cloud data often suffers from incompleteness and irregular distributions. The point cloud completion task aims to recover the complete geometry from partial observations. Despite being inherently ill-posed, this task is critical for downstream 3D understanding applications\cite{liang2018deep,qi2018frustum}.


Recently, numerous deep learning-based approaches have been proposed\cite{chen2023anchorformer,li2023proxyformer,lin2023hyperbolic,tchapmi2019topnet,xiang2021snowflakenet,xie2021style,zhou2022seedformer,wu2024fsc,yuan2018pcn,wei2025pcdreamer}.
Early point cloud completion methods primarily relied on pure 3D network architectures\cite{yuan2018pcn,xie2020grnet,yang2018foldingnet,tchapmi2019topnet,zhou2022seedformer}, such as folding-based decoders or generative models. 
While effective under moderate incompleteness, these methods rely exclusively on the geometric evidence available in partial observations. When large regions are missing, the observed geometry becomes insufficient to constrain the unseen structures, often resulting in ambiguous global shapes and topological errors.
To alleviate this under-constrained inference problem, recent studies\cite{yu2024geoformer,zhu2023svdformer,wei2025pcdreamer} have introduced multi-view 2D--3D representation strategies. These methods project partial point clouds into image-like multi-view representations and employ 2D encoders to extract complementary view-conditioned structural cues for completion.
Moreover, recent studies have further explored multi-modal image guidance and semantic part-level priors for point cloud completion, showing the value of semantic constraints beyond purely geometric decoding\cite{fang2025iaet,zhou2026semantic}. Despite this progress, existing multi-view completion frameworks still face three limitations. 
First, even when coordinate-consistent representations such as Canonical Coordinate Maps are employed, representation-level consistency does not by itself preserve explicit point-to-pixel correspondences for localized feature transfer. Existing interaction modules therefore still rely largely on globally aggregated or learned feature associations, which may weaken the spatial correspondence between observed 3D points and multi-view features.

Second, dense attention for fine-grained point-to-view association
scales poorly with the number of views and spatial tokens, while its
learned weights do not explicitly encode known projective geometry,
limiting geometric interpretability.

Third, heterogeneous features, including coarse structural priors, local point features, and view-conditioned cues, are commonly fused through feature concatenation or learned attention. Although these mechanisms can be input-adaptive, they do not explicitly provide a pointwise mechanism for balancing structural priors against observation-conditioned features. Consequently, the same fusion strategy may behave suboptimally across local regions with different degrees of incompleteness.

In this paper, we propose \textbf{ProjFormer}, a geometrically grounded point cloud completion framework that combines explicit geometric correspondence with learned adaptive feature routing.
Our key idea is twofold: (1) to enforce geometry-consistent 2D--3D interaction through explicit projection, and (2) to perform point-wise adaptive routing between structural priors and observation-driven features.

Following GeoFormer~\cite{yu2024geoformer}, we adopt \textbf{Canonical Coordinate Maps (CCMs)} to encode partial 3D geometry as coordinate-consistent multi-view maps, enabling 2D feature extraction while preserving metric coordinate information. 
The novelty of ProjFormer does not lie in the CCM representation itself, but in how localized CCM features are explicitly associated with observed 3D points and subsequently reused during progressive refinement.
Building on this representation, we design a \textbf{Projective Guided View Attention (PGVA)} module, which uses deterministic projection to retrieve localized features for each observed 3D point and aggregates them through geometry-aware, depth-gated weighting. By sampling only one localized feature from each valid view, PGVA performs point-to-view aggregation with $\mathcal{O}(VN)$ complexity, where $N$ and $V$ denote the numbers of observed points and canonical views, respectively. Compared with the view-level feature fusion used in GeoFormer, PGVA preserves point-aligned CCM features and makes them available during progressive refinement.

Furthermore, we propose a \textbf{Geometry-Aware Routing Refinement} mechanism for progressive refinement. The module decomposes the input features into structural, local
geometric, and CCM-derived view representations, and applies a
point-wise gate to balance structural completion against
observation-conditioned refinement.
Benefiting from the point-aligned tokens produced by PGVA, the routing process enables each point to dynamically balance prior-driven shape completion and observation-conditioned geometric refinement.

\begin{itemize}

\item We propose ProjFormer, a geometry-aware framework for point cloud completion that leverages explicit 2D--3D projection to improve localized point-to-view alignment.

\item We introduce Projective Guided View Attention (PGVA), which uses deterministic projection and geometry-aware, depth-gated weighting to perform point-aligned view aggregation with $\mathcal{O}(VN)$ complexity.

\item We develop a Geometry-Aware Routing Refinement module that performs pointwise adaptive routing between prior-driven structural completion and observation-conditioned geometric refinement.

\item Extensive experiments demonstrate that ProjFormer achieves an average CD$_{\ell_1}$ of 6.34 on PCN while reducing inference latency by approximately 40\% compared with GeoFormer.

\end{itemize}

\section{Related Work}

\subsection{Point Cloud Completion}

Early learning-based point cloud completion methods primarily adopted voxel grids or point-based representations. Voxel- and grid-based approaches~\cite{sharma2016vconv,xie2020grnet} employ 3D CNNs to capture spatial structures, but often suffer from discretization artifacts and high memory consumption. To avoid quantization errors, PCN~\cite{yuan2018pcn} established a point-based encoder--decoder paradigm that predicts a coarse complete shape from a global representation. Subsequent methods improve geometric details through cascaded refinement~\cite{wang2020cascaded} and hierarchical folding~\cite{wen2020point}. SnowflakeNet~\cite{xiang2021snowflakenet} further models parent--child relations during point upsampling through skip-transformers, while SeedFormer~\cite{zhou2022seedformer} introduces \textit{Patch Seeds} to preserve regional information throughout progressive generation.

Following the success of self-attention in general point cloud representation learning~\cite{guo2021pct}, Transformer-based architectures have been introduced into point cloud completion~\cite{yu2021pointr,10232862,wang2024pointattn}. PoinTr~\cite{yu2021pointr} formulates completion as set-to-set translation and employs geometry-aware Transformers to capture long-range dependencies, while subsequent methods improve query generation and feature interaction for more effective shape reconstruction~\cite{10232862,wang2024pointattn}. However, standard dense self-attention incurs quadratic complexity $\mathcal{O}(N^2)$ with respect to the number of point tokens. More importantly, stronger feature aggregation alone cannot fully resolve the ambiguity caused by severely incomplete observations, because the missing geometry is not directly available in the input. This limitation motivates the use of complementary multi-view representations for point cloud completion.

\subsection{Multi-view 2D--3D Fusion for Completion}

Multi-view 2D--3D fusion has become an important paradigm for transferring the dense representation capability of 2D models to sparse 3D point clouds. Representative studies transfer visual knowledge through pixel-to-point distillation and cross-modal contrastive learning~\cite{liu2021learning,sautier2022image,afham2022crosspoint}, while large-scale frameworks such as ULIP, Uni3D, and OpenShape align point clouds with image or vision-language representation spaces~\cite{xue2023ulip,zhou2023uni3d,liu2023openshape}. PointCLIP and PointCLIP V2 further convert point clouds into multi-view image representations to exploit knowledge learned by pretrained vision-language models~\cite{zhang2022pointclip,zhu2023pointclip}. These studies demonstrate that image-based representations can provide complementary dense cues for 3D point cloud understanding.

Motivated by these advances, 2D--3D fusion has also been introduced into point cloud completion. Image-assisted methods combine partial point clouds with paired RGB observations through cross-modal interaction, but require additional image data and calibrated cameras~\cite{fang2025iaet}. To avoid these requirements, self-view completion methods construct image-like representations directly from the partial point cloud. SVDFormer~\cite{zhu2023svdformer} employs self-projected depth maps from multiple views to enhance global shape understanding without external images. GeoFormer~\cite{yu2024geoformer} further introduces Canonical Coordinate Maps (CCMs), which encode canonical 3D coordinates into multi-view maps to preserve geometric consistency across views and enrich point features. More recently, PCDreamer~\cite{wei2025pcdreamer} has explored multi-view diffusion priors to provide additional guidance for recovering missing geometry.

Despite this progress, existing completion frameworks commonly incorporate multi-view evidence through global or view-level descriptors and learned feature interaction. Although effective for enriching shape representations, such mechanisms do not explicitly preserve localized correspondences between individual observed 3D points and their projected image locations. Dense point-to-pixel attention can retain fine-grained interactions but introduces substantial computational overhead as the number of points and image tokens increases. In contrast, ProjFormer follows the self-view setting without requiring external RGB observations. It explicitly projects observed 3D points onto CCM feature maps, retrieves localized features from valid projections, and constructs point-aligned view tokens that are subsequently reused during progressive refinement.

\subsection{Projection-Based Feature Transport}

Projection-based feature transport has also been explored in scene-level 3D perception. Lift-Splat-Shoot and ImVoxelNet lift image features into 3D or BEV space through camera geometry\cite{philion2020lift,rukhovich2022imvoxelnet}, and recent camera-based semantic scene completion methods further use 2D-to-3D transformation for dense occupancy and semantic reasoning\cite{wang2025vlscene,wang2025mixssc,Xue_2026_CVPR}. These works show the value of projection as an alignment prior. 

Unlike dense lifting methods that transform complete image feature maps into voxel or BEV representations, PGVA adopts sparse point-driven sampling for object-level completion. It projects observed 3D points onto canonical multi-view feature maps, retrieves localized features from valid projections, and produces observed-point-aligned tokens that are subsequently reused during progressive refinement.

\begin{figure*}[ht]
    \centering
    \includegraphics[width=1.0\linewidth]{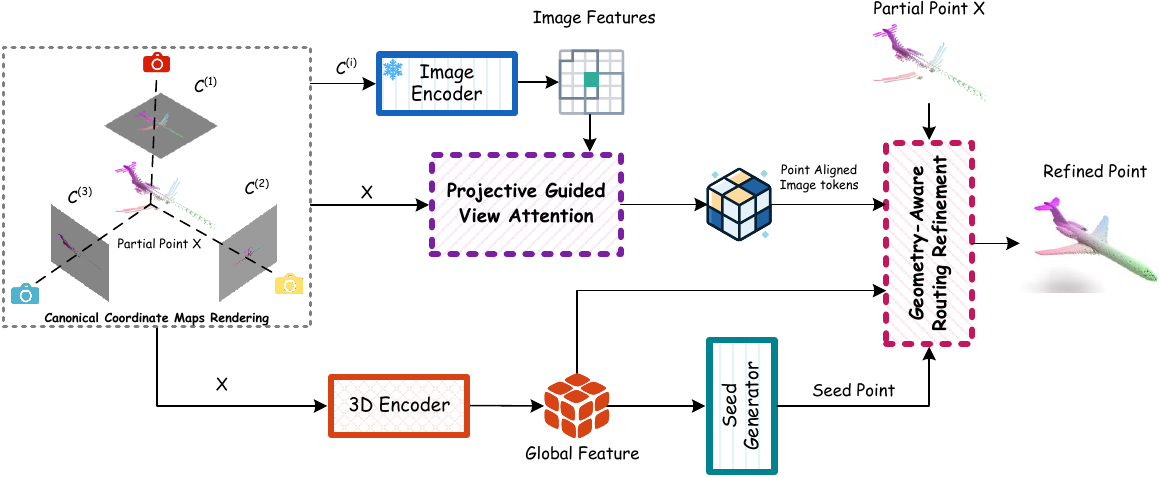}
    \caption{Overview of ProjFormer. The partial point cloud is processed by parallel 3D and CCM-based multi-view branches. PGVA uses explicit projection to construct point-aligned view tokens, which are combined with seed points and the partial input through Geometry-Aware Routing Refinement to generate the completed point cloud.}
    
    \label{fig:pipeline}
\end{figure*}
\section{Methods}

\subsection{Canonical Coordinate Maps}

To leverage mature 2D encoders for 3D point cloud processing, we transform point clouds into multi-view image-grid representations using point-based rasterization.

Given an input point cloud
$\mathcal{X}=\{\mathbf{x}_i\}_{i=1}^{N}$, we first compute its centroid $\boldsymbol{\mu}$ and maximum radial distance $r$. Each point is then mapped to a canonical coordinate:

\begin{equation}
\begin{aligned}
\boldsymbol{\mu}
&=
\frac{1}{N}\sum_{i=1}^{N}\mathbf{x}_i,
\qquad
r
=
\max_i\left\|\mathbf{x}_i-\boldsymbol{\mu}\right\|_2,\\
x'_{i,k}
&=
\frac{x_{i,k}-\mu_k}{2r}
+\frac{1}{2},
\qquad k\in\{1,2,3\}.
\end{aligned}
\end{equation}

This isotropic normalization preserves the aspect ratio and maps the point cloud into a ball of radius $0.5$ centered at $(0.5,0.5,0.5)$, which is contained in $[0,1]^3$. We associate each spatial point with its canonical coordinate as
$\tilde{\mathbf{x}}_i=[\mathbf{x}_i;\mathbf{x}'_i]\in\mathbb{R}^{6}$.

For each predefined viewpoint $v$, let $\mathbf{R}_v$ denote the
world-to-view rotation and let
$\mathbf{t}_v=[0,0,\tau]^\top$ denote the predefined view-space
offset used by the renderer. The corresponding view transformation
and transformed spatial coordinate are
\begin{equation}
\mathbf{P}_v
=
[\mathbf{R}_v\mid-\mathbf{t}_v],
\qquad
\mathbf{x}_i^{(v)}
=
\mathbf{R}_v\mathbf{x}_i-\mathbf{t}_v.
\end{equation}
Only the spatial coordinates are transformed, while the canonical
coordinates $\mathbf{x}'_i$ remain unchanged.

We then apply the fixed perspective mapping used by the CCM
renderer. Let
$\mathbf{x}_i^{(v)}
=
[x_i^{(v)},y_i^{(v)},z_i^{(v)}]^\top$.
For an image of height $H$ and width $W$, the normalized projection
coordinates and the corresponding continuous raster coordinates are
\begin{equation}
\begin{aligned}
\bar{u}_i^{(v)}
&=
\frac{x_i^{(v)}}{z_i^{(v)}}\frac{W}{H},
&
\bar{v}_i^{(v)}
&=
\frac{y_i^{(v)}}{z_i^{(v)}},
\\
\rho_i^{(v)}
&=
\frac{H}{2}
\left(
\bar{u}_i^{(v)}+1
\right),
&
\kappa_i^{(v)}
&=
\frac{W}{2}
\left(
\bar{v}_i^{(v)}+1
\right).
\end{aligned}
\end{equation}

Here, $\rho_i^{(v)}$ and $\kappa_i^{(v)}$ denote the continuous
row and column coordinates, respectively. With the $1\times1$
rasterization footprint used in our implementation, each projected
point is assigned to its nearest discrete pixel
$\mathbf{p}_i^{(v)}$. Points with non-positive depth or projected
locations outside the image boundary are discarded.

Let $\mathcal{I}_{\mathbf{p}}^{(v)}$ denote the set of valid points
assigned to pixel $\mathbf{p}$. Each valid point is assigned an
inverse-depth weight
\begin{equation}
w_i^{(v)}
=
\frac{1}{z_i^{(v)}+\epsilon}.
\end{equation}
When multiple points are assigned to the same pixel, their canonical
coordinates are aggregated as
\begin{equation}
\mathbf{C}^{(v)}(\mathbf{p})
=
\frac{
\displaystyle
\sum_{i\in\mathcal{I}_{\mathbf{p}}^{(v)}}
w_i^{(v)}\mathbf{x}'_i
}{
\displaystyle
\sum_{i\in\mathcal{I}_{\mathbf{p}}^{(v)}}
w_i^{(v)}
},
\qquad
\mathcal{I}_{\mathbf{p}}^{(v)}\neq\varnothing.
\end{equation}
Pixels receiving no valid points are set to zero. The resulting
Canonical Coordinate Maps
$\mathbf{C}^{(v)}\in\mathbb{R}^{3\times H\times W}$
encode the projected canonical coordinates under multiple
predefined viewpoints.

\subsection{Projective Guided View Attention}

Standard cross-modal fusion typically relies on Transformer-based cross-attention between 3D points and 2D pixels, leading to $\mathcal{O}(V \cdot N \cdot HW)$ complexity and ignoring explicit camera geometry. 
Related efficient attention designs in 2D detection and multi-view perception include Deformable DETR and BEVFormer\cite{zhu2020deformable,li2022bevformer}. To address this, we propose Projective Guided View Attention, which leverages projective geometry to establish deterministic point-to-pixel correspondences and achieves efficient multi-view feature aggregation with $\mathcal{O}(V \cdot N)$ complexity.

Instead of learning cross-view attention weights implicitly, PGVA introduces projective geometry as an inductive bias, enabling geometry-consistent and interpretable cross-modal fusion.

\begin{figure}[ht]

    \centering
    \includegraphics[width=1.0\linewidth]{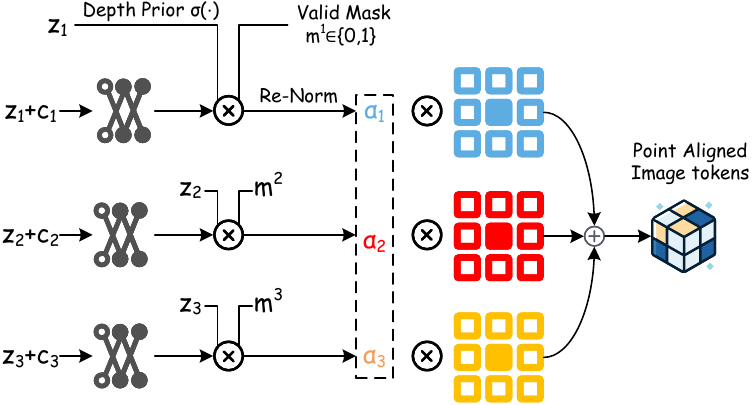}
    \caption{
    Projective Guided View Attention. Each observed 3D point is projected onto multiple CCM views. View features and canonical coordinates are retrieved at valid projected locations and aggregated using geometry-aware, depth-gated weights to produce a point-aligned token.
    }
    
    \label{fig:pgva}
\end{figure}

\noindent\textbf{Projection and Sampling.}
Given a point $\mathbf{x}_i\in\mathcal{X}$ and a view $v$, we reuse
the same deterministic view transformation and fixed perspective
mapping employed in CCM construction:
\begin{equation}
\left(
\rho_i^{(v)},
\kappa_i^{(v)},
z_i^{(v)}
\right)
=
\pi_v(\mathbf{x}_i),
\end{equation}
where $\pi_v(\cdot)$ denotes the projection procedure defined in
the preceding CCM subsection. The corresponding discrete raster
location is denoted by
$\mathbf{p}_i^{(v)}
=
(p_{i,\rho}^{(v)},p_{i,\kappa}^{(v)})$.

\textit{Valid-projection mask generation.}
We set $m_i^{(v)}=1$ if and only if the discretized location lies
within the image boundary and the point has positive view-space
depth:
$0\leq p_{i,\rho}^{(v)}<H$,
$0\leq p_{i,\kappa}^{(v)}<W$, and
$z_i^{(v)}>0$;
otherwise, $m_i^{(v)}=0$.
This mask identifies valid projections rather than explicit
occlusion visibility, since no per-pixel z-buffering is performed.

For each valid projection, we retrieve the image feature
$\mathbf{f}_i^{(v)}$ from $\mathbf{F}^{(v)}$ and the corresponding
canonical coordinate $\mathbf{c}_i^{(v)}$ from $\mathbf{C}^{(v)}$
by direct indexing at the discretized raster location
$\mathbf{p}_i^{(v)}$.

\vspace{0.3em}
\noindent\textbf{Geometry-aware Scoring.}
We evaluate the consistency between the canonical coordinate of the
projected point and the aggregated CCM coordinate retrieved from the
corresponding raster location:
\begin{equation}
e_i^{(v)}
=
\operatorname{MLP}_{\mathrm{score}}
\left(
\left\|
\mathbf{x}'_i-\mathbf{c}_i^{(v)}
\right\|_2,
z_i^{(v)}
\right).
\end{equation}
Both $\mathbf{x}'_i$ and $\mathbf{c}_i^{(v)}$ are expressed in the
same canonical coordinate system, while $z_i^{(v)}$ provides
view-dependent depth information.

\noindent\textbf{Depth-Gated Attention.}
To favor geometrically reliable observations, we incorporate a depth-based gating mechanism with a normalized depth:
\begin{equation}
\bar{z}_i^{(v)}=\operatorname{clip}\!\left(\frac{z_i^{(v)}}{d},\,0,\,1\right),
\end{equation}
where $d$ is the camera distance to the object center used in rendering and $\operatorname{clip}(\cdot,0,1)$ truncates values to $[0,1]$. This normalization is applied in the canonicalized camera frame and stabilizes weights across categories/datasets with different scales. The final attention weight is
\begin{equation}
\alpha_i^{(v)} =
\frac{
\exp(e_i^{(v)}) \cdot \sigma(-\beta\,\bar{z}_i^{(v)}) \cdot m_i^{(v)}
}{
\sum_{v'=1}^{V} \exp(e_i^{(v')}) \cdot
\sigma(-\beta\,\bar{z}_i^{(v')}) \cdot m_i^{(v')}
+ \epsilon
}.
\end{equation}
Here $\sigma(\cdot)$ is the sigmoid function and $\beta>0$
controls the sharpness of depth gating (we set $\beta=1$ by
default). This yields a monotonically decreasing weighting over
normalized depth, approximating a near-field visibility prior while
maintaining cross-category stability.

\noindent\textbf{Feature Aggregation.}
The final point-aligned image feature is obtained as:
\begin{equation}
\boldsymbol{\tau}_i = \sum_{v=1}^{V} \alpha_i^{(v)} \, \mathbf{f}_i^{(v)}.
\end{equation}

Overall, PGVA replaces dense cross-attention with deterministic projection and geometry-guided weighting, resulting in an efficient $\mathcal{O}(V \cdot N)$ formulation while maintaining strong geometric consistency across views.

\begin{figure*}[ht]
    \centering
    \includegraphics[width=1.0\linewidth]{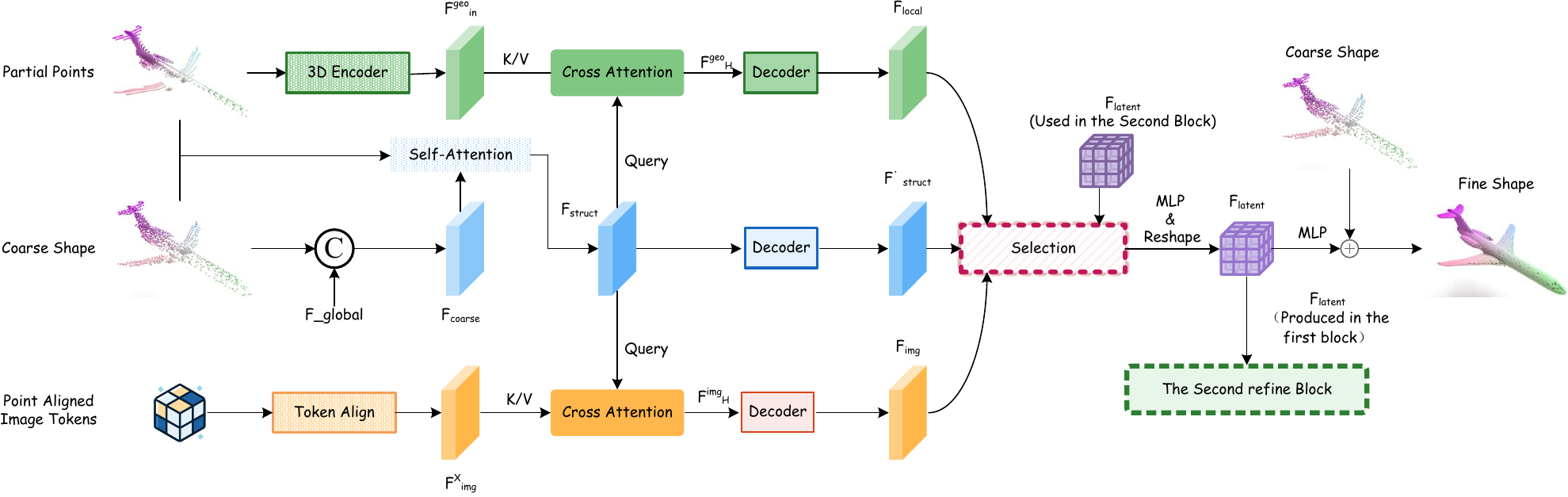}

    \caption{
Geometry-Aware Routing Refinement. Two cascaded refinement blocks adaptively balance structural features against an
observation-conditioned fusion of local geometric and point-aligned view features to predict residual coordinate offsets.
}

    \label{fig:refine_module}
\end{figure*}

\subsection{Geometry-Aware Routing Refinement}
\label{sec:gar_refinement}

While the PGVA module provides geometry-consistent point-aligned image features, effectively integrating multi-source information remains challenging due to the heterogeneous nature of structural priors, local geometry, and 2D semantic priors. Existing methods typically rely on implicit or fixed fusion strategies (e.g., element-wise addition or concatenation), which lack the flexibility to adapt to varying observation conditions. 
For severely missing regions, the model should rely more on structural priors, whereas in well-observed regions it should emphasize observation-conditioned local and view features.
To address this, we propose a Geometry-Aware Routing Refinement module. This component explicitly disentangles information sources and performs point-wise adaptive feature routing. Importantly, since PGVA produces geometry-consistent image tokens, our module can directly leverage them for reliable cross-modal interaction without requiring additional complex alignment learning. The processing logic of Geometry-Aware Routing Refinement is shown in Figure~\ref{fig:refine_module}.

\noindent\textbf{Geometry-aware positional encoding.}
While standard positional embeddings\cite{vaswani2017attention} often rely solely on absolute coordinates, ignoring the relationship between the generated points and the observed input. To inject observation-aware context, we encode the proximity between each coarse point $\widehat{\mathbf{y}}_i$ and the observed input point cloud $\mathcal{X}$ as a geometric confidence signal. Specifically, we compute a normalized point-to-observation distance:
\begin{equation}
d_i = \frac{1}{\sigma_d} \min_{\mathbf{x} \in \mathcal{X}} \|\widehat{\mathbf{y}}_i - \mathbf{x}\|_2,
\end{equation}
where $\sigma_d$ is a scaling factor. This distance $d_i$ is mapped to a sinusoidal embedding $\mathbf{e}_i = \gamma(d_i)$, which serves as a query bias in the attention layers. This encoding implicitly informs the network about the reliability of local geometric observations: points closer to $\mathcal{X}$ are treated as high-confidence anchors, while distant points are flagged for semantic-guided completion.

\smallskip

\noindent\textbf{Multi-source feature decomposition.}
We design three complementary pathways to comprehensively model the shape prior. 
First, to capture global structural dependencies, we apply self-attention on the input feature map $\mathbf{F}$ conditioned on geometric embedding $\mathbf{e}$:
\begin{equation}
\mathbf{F}^{\text{struct}} = \text{SelfAttn}(\mathbf{F}, \mathbf{e}).
\end{equation}
Meanwhile, fine-grained details from the partial input are preserved through a local geometric pathway. Here, we extract multi-scale local features $\mathbf{F}^{\mathcal{X}}_{\text{local}}$ and fuse them with the point features via cross-attention:
\begin{equation}
\mathbf{F}^{\text{local}} = \text{CrossAttn}(\mathbf{F}, \mathbf{F}^{\mathcal{X}}_{\text{local}}).
\end{equation}
Finally, to provide cross-modal image guidance, we exploit the geometry-consistent tokens $\boldsymbol{\tau}$ from PGVA. Given that $\boldsymbol{\tau}$ is already aligned with $\mathcal{X}$, we assign each coarse point the token of its nearest observed neighbor to maintain semantic consistency:
\begin{equation}
j^* = \arg\min_j \|\widehat{\mathbf{y}}_i - \mathbf{x}_j\|_2, \quad
\boldsymbol{\tau}_i = \boldsymbol{\tau}_{j^*}.
\end{equation}
These propagated semantic tokens are then injected into the feature space using another cross-attention mechanism: 
\begin{equation}
\mathbf{F}^{\text{img}} = \text{CrossAttn}(\mathbf{F}, \boldsymbol{\tau}).
\end{equation}

\smallskip

\noindent\textbf{Point-wise adaptive routing.}
Inspired by dynamic routing and feature gating mechanisms in deep learning\cite{hu2018squeeze,chen2020dynamic,shen2023mixture}, we introduce a point-wise gate that dynamically
balances the structural pathway against an observation-conditioned pathway. The latter is constructed by equally combining the local geometric feature and the point-aligned view feature. A routing gate $g_i\in(0,1)$ is predicted for each point by aggregating the decomposed features, the global context vector $\mathbf{f}_g$, and the previous-stage feature $\mathbf{F}_{\mathrm{pre}}$:
\begin{equation}
\mathbf{g}_i = \sigma\Big(\mathrm{MLP}\big(\text{Concat}[\mathbf{F}^{\text{struct}}_i, \mathbf{F}^{\text{local}}_i, \mathbf{F}^{\text{img}}_i, \mathbf{f}_g, \mathbf{F}_{\text{pre}, i}]\big)\Big),
\end{equation}
where $\sigma$ denotes the sigmoid function. The final fused feature $\mathbf{F}_i$ is computed as:
\begin{equation}
\mathbf{F}_i =
\mathbf{g}_i \cdot \mathbf{F}^{\text{struct}}_i
+
(1 - \mathbf{g}_i) \cdot \tfrac{1}{2}
(\mathbf{F}^{\text{local}}_i + \mathbf{F}^{\text{img}}_i).
\end{equation}

Thus, the gate favors structural priors in missing regions and
observation-guided features on observed surfaces.

\smallskip
\noindent\textbf{Progressive refinement.}
The fused features are decoded into residual coordinate offsets and applied to upsampled points to generate the refined geometry. We adopt a cascaded refinement strategy where the latent features $\mathbf{F}_L$ from one stage are propagated to the next (denoted as $\mathbf{F}_{\text{pre}}$), ensuring continuity in feature learning:
\begin{equation}
\widehat{\mathcal{Y}}^{(s)} =
\text{Upsample}(\widehat{\mathcal{Y}}^{(s-1)}, r_s)
+
\text{MLP}_{\text{dec}}(\mathbf{F}^{(s)}),
\end{equation}
where $s$ and $r_s$ denote the stage index and upsampling ratio, respectively; the cascade progressively recovers coarse topology and fine-grained geometry.


\noindent\textbf{Training objective.}
We supervise the coarse and two refined predictions with
HyperCD~\cite{lin2023hyperbolic}. Let
$\mathcal{Y}^{(0)}$ and $\mathcal{Y}^{(1)}$ be ground-truth point
sets downsampled by farthest point sampling to match
$\widehat{\mathcal{Y}}^{(0)}$ and $\widehat{\mathcal{Y}}^{(1)}$,
respectively, and let $\mathcal{Y}^{(2)}=\mathcal{Y}$. We optimize
\begin{equation}
\mathcal{L}_{\mathrm{train}}
=
10^{3}\!\left[
\sum_{s=0}^{2}
\mathcal{L}_{\mathrm{HCD}}
\bigl(\widehat{\mathcal{Y}}^{(s)},\mathcal{Y}^{(s)}\bigr)
+
\mathcal{L}_{\mathrm{HCD}}^{\rightarrow}
\bigl(\mathcal{X},\widehat{\mathcal{Y}}^{(2)}\bigr)
\right],
\end{equation}
where the one-sided term preserves the observed input geometry.

\begin{table*}[!htbp]
\centering
\small 
\setlength{\tabcolsep}{6pt} 
\caption{Quantitative comparison of point-cloud completion results on the PCN dataset measured by the $\mathrm{CD}_{\ell_1}$ distance ($\times 10^{3}$). }
\begin{tabular}{lcccccccccc} 
\toprule
Method & Venue & Average & Plane & Cabinet & Car & Chair & Lamp & Sofa & Table & Vessel \\
\midrule
PCN        & 3DV'18   & 10.60 & 6.09 & 11.55 & 9.51 & 12.33 & 12.57 & 12.87 & 9.87  & 9.98 \\
PoinTr     & ICCV'21  & 8.38 & 4.75 &10.47  & 8.68 & 9.39 & 7.75 & 10.93 & 7.78 & 7.29 \\
SeedFormer & ECCV'22  &  6.74 & 3.85 &  9.05 & 8.06 &  7.06 &  5.21 & 8.85 & 6.05  & 5.85 \\
SVDFormer  & ICCV'23  & 6.55 & 3.64  & 8.81 & 7.44 & 6.93 & 5.32 & 8.60 & 5.86 & 5.80\\
CRA-PCN    & AAAI'24  & 6.56 & 3.62 & 8.77 & 7.00 & 6.92 & 5.46 & 8.59 & 6.27 & 5.86 \\
ODGNet     & AAAI'24  &   6.50 &  3.77 & 8.77 &  7.56 &  6.84 &  5.09 & 8.47 & 5.84 & 5.66 \\
FSC        & CVPR'24  &  7.02 & 4.07 &  9.12 & 8.10 &  7.21 &  5.88 & 9.30 & 6.26  & 6.25 \\
GeoFormer  & ACMMM'24 &  6.42 & 3.60 &  8.69 & 7.46 &  6.71 &  5.15 & 8.28 & 5.84 & 5.63 \\
PCDreamer  & CVPR'25  & 6.52 & 3.51  & 8.62 & \textbf{6.92} & 6.91 & 5.66 & \textbf{8.31} & 6.27 & 5.90\\
\midrule 
Ours       & ACMMM'26      & \textbf{6.34} & \textbf{3.58} & \textbf{8.61} & 7.37 & \textbf{6.62} & \textbf{4.95} & 8.32 & \textbf{5.76} & \textbf{5.51} \\
\bottomrule
\end{tabular}

\label{tab:pcn_results_2048}
\end{table*}

\begin{table*}[t]
\centering

\caption{Comparison with state-of-the-art methods on the
ShapeNet-55/34 benchmarks using $\mathrm{CD}_{\ell_2}\times10^3$.
Lower is better ($\downarrow$).}

\begin{tabular}{l|ccc|ccc|ccc}
\hline
\multirow{2}{*}{Method} & \multicolumn{3}{c|}{ShapeNet 55 dataset} & \multicolumn{3}{c|}{34 seen categories} & \multicolumn{3}{c}{21 unseen categories} \\
 & CD-S ($\downarrow$) & CD-M ($\downarrow$) & CD-H ($\downarrow$) & CD-S ($\downarrow$) & CD-M ($\downarrow$) & CD-H ($\downarrow$) & CD-S ($\downarrow$) & CD-M ($\downarrow$) & CD-H ($\downarrow$) \\
\hline
SeedFormer & 0.50 & 0.77 & 1.49 & 0.48 & 0.70 & 1.30 & 0.61 & 1.07 & 2.35 \\
ODGNet & 0.47 & 0.70 & 1.32 & 0.44 & 0.64 & 1.14 & 0.59 & 1.01 & 2.26 \\
SVDFormer & 0.48 & 0.70 & 1.30 & 0.46 & 0.64 & 1.13 & 0.61 & 1.05 & 2.19 \\

CRA-PCN & 0.48 & 0.71 & 1.37 & 0.45 & 0.65 & 1.18 & 0.55 & 0.97 & 2.19 \\
GeoFormer & 0.41 & 0.64 & 1.25 & 0.39 & 0.57 & \textbf{1.05} & 0.55 & 0.99 & 2.15 \\
\hline
\textbf{Ours} & \textbf{0.39} & \textbf{0.58} & \textbf{1.24} & \textbf{0.38} & \textbf{0.54} & 1.10 & \textbf{0.44} & \textbf{0.81} & \textbf{1.93} \\
\hline
\end{tabular}

\label{tab:comparison_srn}

\end{table*}

\begin{figure*}[ht]
    \centering
    \includegraphics[width=1.0\linewidth]{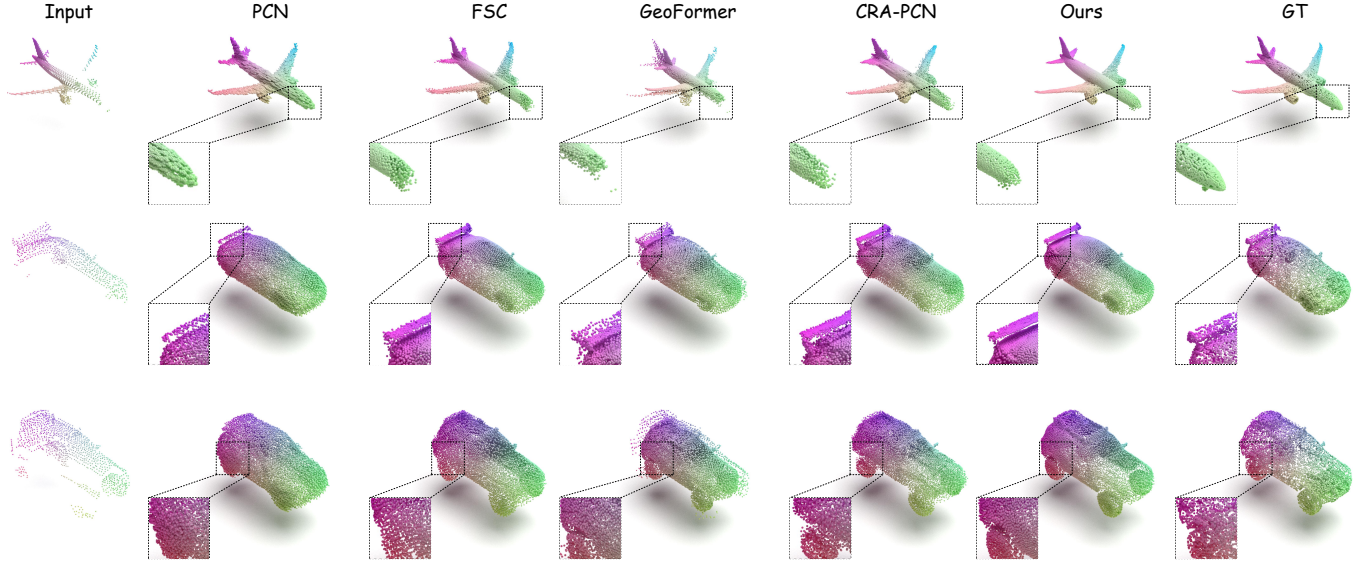}
    \caption{
    Visual comparisons on the PCN dataset.
    }
    
    \label{fig:visualize}
\end{figure*}

\section{Experiment}


\subsection{Experimental Setup}
\textbf{Datasets.} We evaluate ProjFormer on PCN\cite{yuan2018pcn}, ShapeNet-55/34\cite{shapenet2015}, and KITTI\cite{geiger2013vision} following standard protocols~\cite{yuan2018pcn,yu2021pointr,zhou2022seedformer}. 

PCN contains 8 categories with partial inputs of 2{,}048 points and ground truth of 16{,}384 points, synthesized from 8 viewpoints; the dataset totals 30{,}974 shapes split into training (28{,}974), validation (800), and testing (1200). 
To simulate realistic inputs, we sample points from meshes, back-project 2.5D depth images to obtain 2048-point partials.

ShapeNet-55/34 provide broad category coverage: ShapeNet-55 comprises 41952 training and 10518 testing shapes across 55 categories; ShapeNet-34 includes 46765 training shapes and evaluates both seen (34 categories; 3400 shapes) and unseen (21 categories; 2305 shapes) categories. During testing, three difficulty levels—Simple, Moderate, Hard—are defined by incomplete ratios (25\%, 50\%, 75\%). 

Since the previous two datasets are synthetic, generated from CAD models or meshes, they may not fully reflect the characteristics of real scanned point clouds. To address this gap, we further incorporate the KITTI dataset~\cite{geiger2013vision}, which is collected from an autonomous driving platform and serves as a challenging real-world benchmark. 

\textbf{Metrics.}
We report Chamfer Distance (CD) and F-Score following established
practice. On PCN, we use $\mathrm{CD}_{\ell_1}$:
\begin{equation*}
\mathrm{CD}_{\ell_1}(\hat{P}, P)
=
\frac{1}{2}
\left(
\frac{1}{|\hat{P}|}
\sum_{x \in \hat{P}}
\min_{y \in P}
\|x-y\|_2
+
\frac{1}{|P|}
\sum_{y \in P}
\min_{x \in \hat{P}}
\|y-x\|_2
\right),
\end{equation*}
where $\hat{P}$ and $P$ denote the completed and ground-truth point
sets, respectively. Here, $\mathrm{CD}_{\ell_1}$ refers to the
unsquared Euclidean Chamfer distance. Following the PCN benchmark
convention, the reported values are multiplied by $10^3$.

For KITTI, where complete ground truth is unavailable, we adopt a distribution-level metric, Minimum Matching Distance (MMD), to assess realism of completions. Given predicted KITTI completions $\mathcal{P}=\{\hat{\mathbf{X}}_i\}_{i=1}^{N}$ and a reference set of complete car shapes $\mathcal{G}=\{\mathbf{Y}_j\}_{j=1}^{M}$ (formed by the union of train/val/test car samples in PCN/ShapeNet-Cars), MMD is defined as
\begin{equation}
\mathrm{MMD}(\mathcal{P},\mathcal{G})
=
\frac{1}{N}\sum_{i=1}^{N}\min_{1\le j\le M}
d_{\mathrm{CD}}(\hat{\mathbf{X}}_i,\mathbf{Y}_j),
\end{equation}
where $d_{\mathrm{CD}}$ denotes the bidirectional Chamfer Distance (L2)
Lower MMD indicates that predicted KITTI shapes are closer to the manifold of realistic complete car geometries.

\subsection{Main Results}

\begin{table*}[!t] 
\centering
\renewcommand{\arraystretch}{1.1} 
\setlength{\tabcolsep}{5pt} 

\caption{Ablation study on 4 different model components.}
\begin{tabular}{c l c c c c c c c}
\toprule
\textbf{No.} & 
\textbf{Model Configuration} & 
\textbf{CCM} & 
\textbf{Point-aligned token} & 
\textbf{Depth Gating} & 
\textbf{PGVA} & 
\textbf{$\mathrm{CD}_{\ell_1}$ ($\downarrow$)} & 
\textbf{$\mathrm{CD}_{\ell_2}$ ($\downarrow$)} & 
\textbf{F-Score@1\% (\%) ($\uparrow$)} \\
\midrule

1 & Full w/o CCM               &            & \cmark & \cmark & \cmark & 6.785 & 2.0466 & 82.1679 \\
2 & Full w/o Point-aligned     & \cmark &            & \cmark & \cmark & 6.448  & 1.8129 & 84.1216 \\
3 & Full w/o depth gating      & \cmark & \cmark &            & \cmark & 6.384 & \textbf{1.7599} & 84.4839 \\
4 & Full w/o PGVA              & \cmark &  &  &            & 6.493 & 1.8468 & 83.6134 \\
\midrule

5 & \textbf{Full Model}        & \cmark & \cmark & \cmark & \cmark & \textbf{6.3446} & 1.7905 & \textbf{84.9282} \\
\bottomrule
\end{tabular}

\label{tab:ablation_4_components}
\end{table*}

\textbf{PCN.} We compare ProjFormer against recent state-of-the-art methods, categorized into purely 3D-based approaches (SeedFormer~\cite{zhou2022seedformer}, CRA-PCN~\cite{rong2024cra}) and 2D--3D cross-modal frameworks (e.g., GeoFormer~\cite{yu2024geoformer}, SVDFormer~\cite{zhu2023svdformer}). The latter group leverages multi-view projections to exploit semantic information similar to our approach, providing a direct benchmark for our cross-modal design. As shown in Figure~\ref{fig:visualize}, ProjFormer produces more
uniform completions with fewer structural artifacts and outliers.


\textbf{ShapeNet-55/34.} We extend the experiments to ShapeNet-55 and the ShapeNet-34, the two settings cover the 55-category ShapeNet completion benchmark.

Following the protocol established in previous works \cite{rong2024cra,yu2024geoformer,zhou2022seedformer,ODGNet}, partial observations are synthesized during the training stage. 

During training, random viewpoints are selected, and the $n\%$ farthest points from these viewpoints are discarded to simulate partial scans. 
At test time, the viewpoints are fixed, and $n$ is systematically varied to three difficulty levels: $25\%$ (simple), $50\%$ (moderate), and $75\%$ (hard).

Our experimental evaluation results on ShapeNet55/34 are presented in Table~\ref{tab:comparison_srn} . We compare our method with existing point cloud completion baselines, using  $\mathrm{CD}_{\ell_2}$ as evaluation metrics, the reported $\mathrm{CD}_{\ell_2}$ values are multiplied by $10^3$.

\textbf{KITTI.} Since the point clouds in KITTI are obtained from real LiDAR scans and contain no ground-truth shapes, direct per-sample reconstruction error cannot be computed. We therefore evaluate realism using the Minimum Matching Distance (MMD) defined in Experimental Setup, where lower values indicate closer alignment to realistic complete car geometries. We additionally report Fidelity Distance (FD), computed as the one-sided distance from the observed input points to the completed point cloud, to measure preservation of the input geometry.

\begin{figure}[t]
    \centering
    \includegraphics[width=1.0\linewidth]{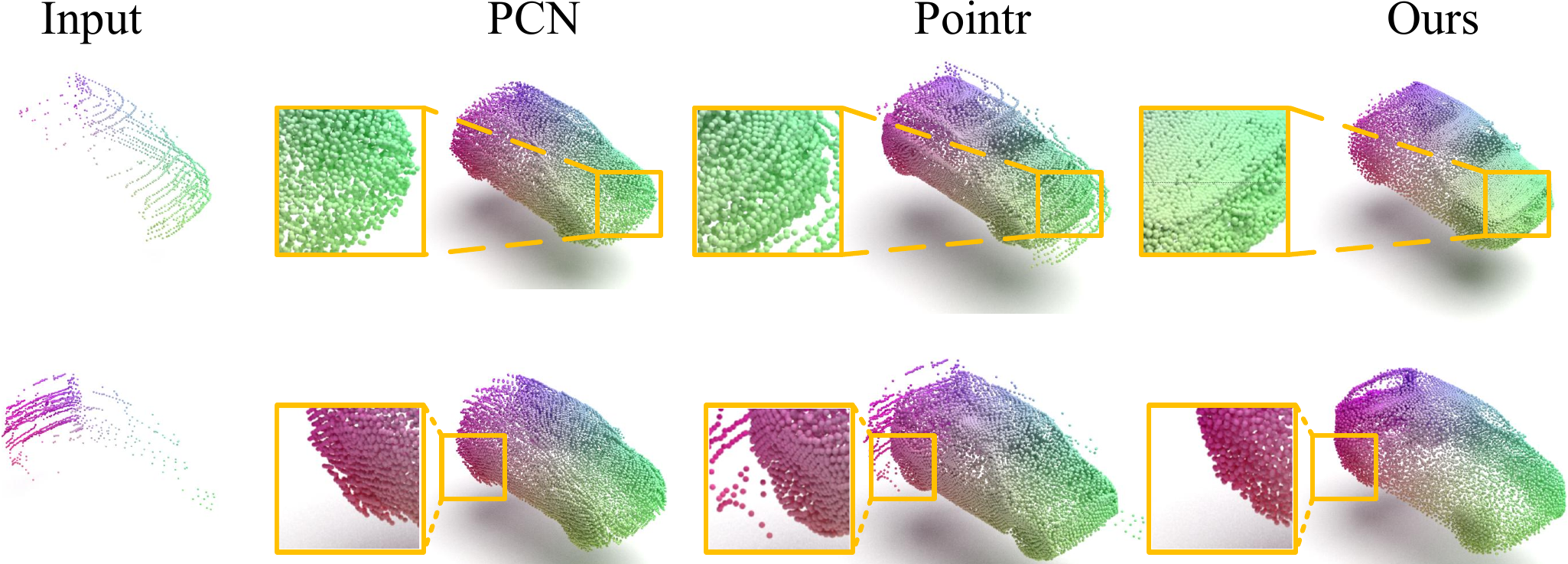}
    \caption{Qualitative completion results on the KITTI dataset.}
    \label{fig:KITTI_Compare}
\end{figure}

The point cloud completion results on KITTI are illustrated in Fig.~\ref{fig:KITTI_Compare}. The results of the quantitative experiments are shown in Table ~\ref{tab:real_world_results}.
Our method achieves the lowest MMD, indicating that its predictions are closer to the reference distribution of complete car shapes. However, the higher FD reveals a trade-off between shape plausibility and preservation of the observed input geometry.

\textbf{Computational Efficiency.} Beyond reconstruction accuracy, we evaluate the inference efficiency of ProjFormer against mainstream methods in Table~\ref{tab:efficiency}. Notably, the reported inference time (15.85 ms) natively includes the overhead of rendering 2D images from partial inputs. While sharing a 2D--3D fusion paradigm with GeoFormer and SVDFormer, our architecture bypasses their computationally heavy cross-view attention mechanisms.

\begin{table}[t]
\centering

\caption{Quantitative comparison on KITTI}
\begin{tabular}{l|c|c|c}
\hline
Methods & PoinTr & SeedFormer  & Ours \\ \hline
MMD $\downarrow$ &8.21 & 1.09 & \textbf{0.48} \\
FD $\downarrow$ & 0.00 & \textbf{1.45} & 1.97 \\
\hline
\end{tabular}
\label{tab:real_world_results}

\end{table}

\begin{table}[!t]
\centering
\small
\setlength{\tabcolsep}{5pt}
\caption{Comparison of completion accuracy and inference speed on the PCN dataset.}
\begin{tabular}{llcc}
\toprule
Method & Venue & \makecell{$\mathrm{CD}_{\ell_1}$} & \makecell{Infer Time(ms)} \\
\midrule
CRA-PCN    & AAAI'24  & 6.56 & 67.61 \\
ODGNet     & AAAI'24  & 6.50 & 45.39 \\
SVDFormer  & ICCV'23  & 6.55 & 69.40 \\
FSC        & CVPR'24  & 7.02 & 26.97 \\
GeoFormer  & ACMMM'24 & 6.42 & 26.37 \\
\midrule
Ours       & ACMMM'26      & \textbf{6.34} & \textbf{15.85} \\
\bottomrule
\end{tabular}
\label{tab:efficiency}
\end{table}

\subsection{Ablation Studies}
Table~\ref{tab:ablation_4_components} evaluates the contributions of the main components on PCN. In the Full w/o CCM variant, we replace the three-channel
CCMs with the standard self-projected depth maps adopted by
SVDFormer~\cite{zhu2023svdformer}, retaining only depth information while removing the canonical coordinate cues.Removing the CCM coordinate cues increases CD$_{\ell_1}$ from 6.3446 to 6.785, confirming the importance of coordinate-consistent multi-view representations. Removing the point-aligned tokens increases CD$_{\ell_1}$ to 6.448, showing that reusing localized CCM features during refinement benefits geometric reconstruction. Disabling depth gating slightly increases CD$_{\ell_1}$ to 6.384 and decreases the F-Score to 84.48, although CD$_{\ell_2}$ exhibits a minor fluctuation. Finally, replacing PGVA with global image pooling increases CD$_{\ell_1}$ to 6.493, demonstrating that localized observed-point-to-view aggregation is more effective than global feature fusion. Additional ablations on the number and configuration of canonical views are provided in the supplementary material.

\section{Conclusion}
We presented ProjFormer, an efficient cross-modal framework for point cloud completion. PGVA establishes geometry-consistent point-to-view correspondences, while Geometry-Aware Routing Refinement adaptively
combines structural, local, and image-derived features during
progressive reconstruction. Experiments demonstrate competitive
completion accuracy with substantially lower inference latency than comparable cross-modal methods, highlighting a favorable balance between reconstruction quality and computational efficiency.

\begin{acks}
This work was supported by the National Natural Science Foundation of China
(Grant No. U25A20421) and the Project of Yuelushan Center for Industrial
Innovation (Grant No. 2025YCII0225).
\end{acks}

\bibliographystyle{ACM-Reference-Format}
\bibliography{projformer-references}

\end{document}